\documentclass[runningheads]{llncs}

\usepackage[T1]{fontenc}
\usepackage{lmodern}
\usepackage{amsmath,amssymb}
\usepackage{graphicx}
\usepackage{mdframed}
\usepackage{xcolor}
\usepackage[T1]{fontenc}
\usepackage{booktabs}    
\usepackage{tabularx}    
\usepackage{array}    
\usepackage{graphicx}
\usepackage{tabularx}

\usepackage[ruled,vlined]{algorithm2e}

\usepackage{listings}

\usepackage{enumitem}
\usepackage{amsmath}
\usepackage{amssymb}

\begin{document}

\title{CutClean: Neural Network Pruning for Privacy-Preserving Inference}

\author{Leonardo Magliolo\inst{1} \and
Vito Paolo Pastore\inst{2,3} \and Giuseppe Valenzise\inst{4} \and 
Enzo Tartaglione\inst{1}}
\authorrunning{L. Magliolo et al.}
%
\institute{LTCI, Télécom Paris, Institut Polytechnique de Paris, France \and MaLGa-DIBRIS, University of Genoa, Italy \and Istituto Italiano di Tecnologia, Italy \and Universite Paris-Saclay, CNRS, CentraleSupélec, L2S, France \\
\email{\{leonardo.magliolo, enzo.tartaglione\}@telecom-paris.fr} \\
\email{giuseppe.valenzise@l2s.centralesupelec.fr} \\
\email{vito.paolo.pastore@unige.it}}
\maketitle

\begin{abstract}
Neural networks are increasingly deployed in high-stakes applications with growing privacy leakage concerns. 
We show that this privacy leakage can occur even in the absence of representation imbalances that lead to 
traditional dataset biases. This poses significant privacy risks when deploying models that process sensitive 
attributes. In this context, we propose CutClean, a privacy-aware pruning method that allows to reduce privacy information flow through the network, while increasing its sparsity. Our approach employs auxiliary linear privacy heads placed at each network's block to quantify information 
leakage, and further applies increasing levels of sparsity to remove the private attribute leakage, measured in terms of the accuracy of the privacy head attached to the last block.
Experiments on synthetic and real-world datasets demonstrate that our approach effectively minimizes private information flow while achieving high sparsity rates and preserving classification target accuracy.\footnote{This work has been accepted for publication at the International Conference on Pattern Recognition 2026.}

\end{abstract}

\section{Introduction}

\label{sec:1_introduction}

Neural networks are increasingly deployed in high-stakes domains such as healthcare, finance, biometric authentication, and social services, where automated decisions can have significant societal and individual consequences. In response to these risks, recent regulatory frameworks—most notably the European Union’s AI Act~\cite{act2024eu}—have placed strong emphasis on privacy protection, data minimization, and safeguards against unintended uses of sensitive information~\cite{selbst2019fairness}. Despite these advances, modern neural networks remain prone to side leakage: even when models are trained without explicit access to sensitive attributes, their internal representations can encode private information that may later be extracted by adversaries, auxiliary classifiers, or downstream tasks~\cite{fredrikson2015model,shokri2017membership}. This misalignment between regulatory intent and model behavior poses substantial privacy risks, particularly in real-world deployments where access to trained models or intermediate representations cannot be fully controlled. 

Existing work on privacy leakage in neural networks has largely focused on biases induced by dataset imbalances or correlations between sensitive and target attributes. In these settings, sensitive information emerges as a byproduct of spurious correlations, and mitigation strategies often rely on dataset balancing, fairness constraints, or adversarial representation learning~\cite{sagawa2020investigation}. However, privacy leakage can persist even in the absence of such imbalances~\cite{ravfogel2020null,locatello2019challenging}: neural networks may encode sensitive attributes simply because they are predictive of intermediate features or useful for representation learning, rather than due to explicit bias in the training data. This observation challenges the assumption that removing dataset bias alone is sufficient to prevent privacy leakage, and highlights the need for model-centric privacy interventions. 

In parallel, the machine learning community has explored post-hoc techniques to modify trained networks in order to improve efficiency, robustness, or interpretability. Among these, network pruning~\cite{han2015learning,liao2023can} has emerged as a powerful tool to reduce model size and computational cost by removing redundant parameters or structures. Structured pruning methods, in particular, remove entire channels, filters, or blocks, enabling deployment-friendly sparsity patterns~\cite{bragagnolo2021role}. Despite their success in efficiency-oriented settings, the potential of pruning as a privacy-preserving mechanism remains underexplored. Intuitively, removing parts of a network that encode sensitive information could reduce privacy leakage~\cite{voita2019analyzing}, but doing so in a principled and measurable way remains an open challenge.

In this work, we propose CutClean, a privacy-aware structured pruning framework designed to explicitly reduce private attribute leakage in neural networks. Our approach introduces auxiliary linear privacy heads attached to intermediate blocks of a network, allowing us to quantify how much private information is present at different depths. Rather than relying on dataset-level assumptions, CutClean directly measures privacy leakage through the accuracy of these privacy heads. We then progressively apply structured pruning to remove components that contribute most to private attribute predictability, while monitoring both target task performance and privacy leakage. We empirically validate CutClean on both synthetic and real-world datasets, demonstrating that it achieves high sparsity rates and substantial reductions in private attribute predictability without sacrificing classification performance.
We summarize our contributions as follows:
\begin{itemize}
    \item We introduce CutClean, the first structured pruning framework explicitly designed to reduce private attribute leakage in trained neural networks, positioning pruning as a privacy-preserving intervention rather than a compression or robustness tool (Sec.~\ref{sec:approach}).
    \item We show that significant reductions in private attribute leakage can be achieved without adversarial minimax optimization, improving stability and post-deployment applicability (Sec.~\ref{sec:results}).
    \item We demonstrate that CutClean achieves high structured sparsity while preserving target task accuracy, revealing an unexpected compatibility between privacy leakage reduction and efficient model deployment (Sec.~\ref{sec:ablations}).
\end{itemize}
\section{Related works}
\label{sec:2_related_works}
\textbf{Privacy Leakage and Adversarial Attribute Inference.} A growing body of work has shown that neural network representations can leak sensitive information even when protected attributes are not explicitly used during training~\cite{fredrikson2015model,shokri2017membership}. Attribute inference attacks and auxiliary classifier attacks demonstrate that sensitive attributes such as gender, ethnicity, or health status can often be recovered from learned features with high accuracy~\cite{fredrikson2015model,song2019auditing,liu2022membership}. These findings have motivated the use of adversarial classifiers, often referred to as privacy heads or attribute classifiers~\cite{irene}, to quantify or exploit privacy leakage in intermediate representations~\cite{edwards2015censoring,ganin2016domain}. In this setting, privacy risk is typically measured by the accuracy of an attacker trained to predict a sensitive attribute from a model’s internal features.

\noindent \textbf{Debiasing and Privacy.} Beyond adversarial training, several approaches have investigated post-hoc techniques for mitigating bias and privacy leakage, including feature projection, representation obfuscation, and fine-tuning with regularization constraints~\cite{ravfogel2020null,belrose2023leace}. These methods often focus on fairness metrics or demographic parity, rather than explicitly quantifying and minimizing information leakage~\cite{zhang2018mitigating,beutel2017data}. Few works consider structured model modification as a mechanism to control privacy risk after training, particularly without modifying the original loss function or requiring repeated adversarial retraining.

\noindent \textbf{Pruning in Neural Networks.} Network pruning aims to reduce model complexity by removing redundant parameters, channels, or structural components~\cite{han2015learning,liao2023can,ali2024trimming}. While pruning has traditionally been studied in the context of model compression, acceleration, and robustness, recent work has begun to explore its impact on representation properties~\cite{stahl2025distillation,cassano2025does}. However, most existing pruning strategies are agnostic to privacy or bias considerations and rely solely on task performance or weight magnitude as pruning criteria.

CutClean differs from prior work by leveraging pruning to privacy leakage reduction, guided by explicit measures of sensitive attribute predictability. Rather than relying on adversarial minimax optimization, our approach uses auxiliary linear privacy heads solely as diagnostic tools to quantify information flow at different network depths. By progressively pruning network components that contribute to private attribute leakage, CutClean avoids the instability and retraining overhead of adversarial debiasing while enabling deployment-friendly sparsity. This positions structured pruning as a practical and effective post-hoc privacy-preserving mechanism, complementing existing training-time approaches to debiasing and privacy protection. The success of CutClean is prompted by adversarial robustness works showing that properly pruned architectures can be robust to adversarial attacks~\cite{ye2019adversarial,madaan2020adversarial}: differently from these, CutClean aims at removing specific attributes that can be in-distribution at training and are not necessarily outliers.
\section{Method}
\label{sec:3_method}

\subsection{Problem Formulation}
In this section, we define the main concepts and objectives of this work. We want a neural network $f: \mathcal{X} \rightarrow \mathcal{Y}$ to learn the mapping between inputs $x$ and output labels $\hat{y}$ while preventing the model from conveying information about a secondary private label $\hat{z} \in \mathcal{Z}$. The network $f$ is composed of successive blocks $\{b_{i}\}_{i=1}^{B}$, with intermediate output described as: 
\begin{equation}
    \forall i \in [1,B], \quad h_i = b_i(h_{i-1}),
\end{equation}
with $h_0 = \mathbf{x}$ and $\hat{\mathbf{y}} = g^{c}(h_B)$, where $g^{c}$ is the softmaxed linear used as the task-specific classification head. 

We estimate the amount of information conveyed by each intermediate output throughout the network attaching one auxiliary linear layer to the output of \emph{each} block,  acting as a \textit{privacy head}:
\begin{equation}
    \forall i \in [1,B], \quad \hat{z}_i = g^{p}_{i}(h_i).
\end{equation}

In the naive setup, the main model is trained by minimization of a cross-entropy loss $\mathcal{L}_y(\hat{y})$ and each privacy head $g^{p}_{i}$ is trained by minimization of the cross-entropy loss $\mathcal{L}_z(\hat{z}_i)$ whose error is not propagated to the backbone $f$. The performance of the privacy heads can be seen as a measure of the private information that can be extracted from the corresponding block outputs.

We propose to compute a series of pruning masks $\{m_i\}_{i=1}^B$ applied to network channels in each corresponding block, aiming to reach the best trade-off between performance on the main task and sparsity in the pruned network, while minimizing the amount of private information extractable from the network representations, \textit{i.e.} decreasing the performance of the privacy heads close to random guess.
The definition of a block depends on the specific architecture, as described in Sec.~\ref{subsec:imp}.

\subsection{Proposed Approach}
\label{sec:approach}
\begin{figure}[t]
    \centering
    \includegraphics[width=\linewidth]{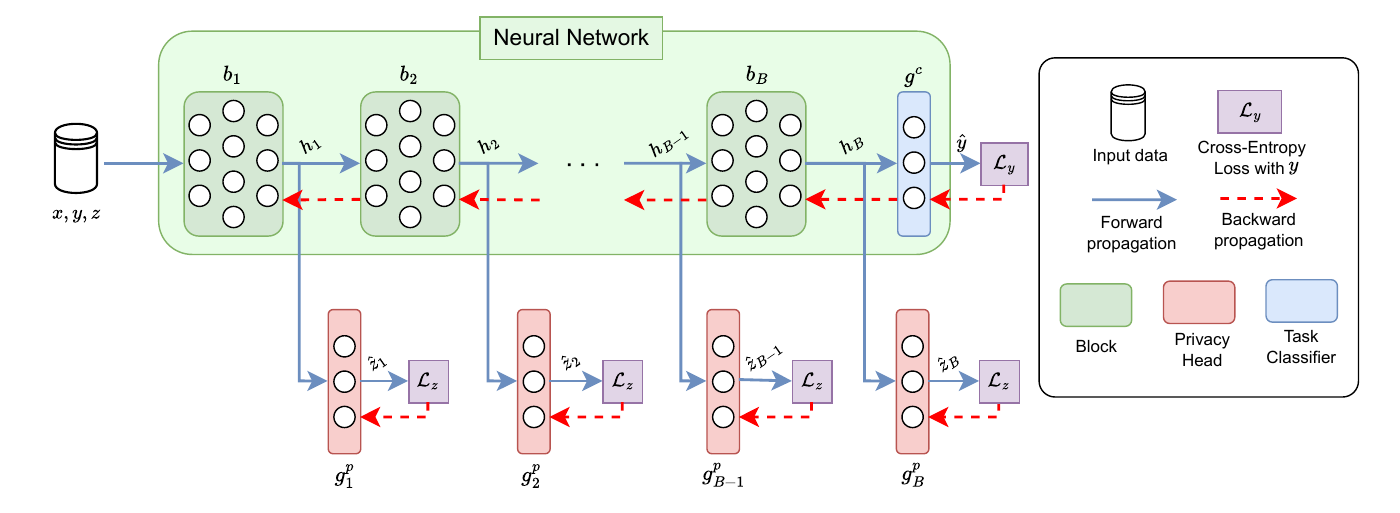}
    \caption{Schematic overview of CutClean.}
    \label{fig:method-setup}
\end{figure}
The goal of this work is to prune the network under the constraint of minimizing the private information flowing through the network representations. To do so, we add adversarial terms in the optimization objective to reduce the performance of optimally trained privacy heads attached to the outputs of all blocks. We measure the private information flow as the accuracy of the privacy head attached to the last network's block. The proposed privacy-aware pruning strategy, that we refer to as CutClean, is schematically represented in Fig.~\ref{fig:method-setup}, while its pseudocode description is reported in the Supplementary material. First, we pre-train our network applying an alternating training procedure to obtain a privacy-aware backbone, that we refer to as online batch Mutual Information (MI)-aware training. Then, we perform a custom pruning scheme exploring different candidate sparsity levels with a grid search over a fixed grid ($s \in \left[0,1\right]$). To control the trade-off between sparsity and private information flow, we monitor the validation accuracy of the last privacy head. We define a sparsity level as admissible, if the last privacy head accuracy, after fine-tuning, is lower or equal to a  defined threshold $P^{\text{threshold}}$. We fine-tune each pruned model corresponding to an admissible sparsity level for a small fixed number of epochs, using the alternating MI-aware training procedure employed for the pre-training. Among the admissible pruned models, the one performing best on the validation set is selected and employed for evaluation on the test set.
We will now provide more details on the main components of our pipeline, including the online batch MI-aware training (Sec.~\ref{sec:MI}) and the designed pruning strategy (Sec.~\ref{sec:pruning}). 

\subsection{Online Batch MI-aware Training}
\label{sec:MI}
Naively, we could employ an optimization function such as
\begin{equation}
    \label{eq:naive_optimization}
    \mathcal{J}=\mathcal{L}_y(\hat{y}) - \sum_{i=1}^B \mathcal{L}_z(\hat{z}_i).
\end{equation}

However, as shown in \cite{irene}, such an optimization goal could lead to degenerate solutions in which the privacy head systematically misclassifies the private attributes with high confidence.
This behavior still requires the backbone to encode the private attribute reliably (the head must infer the correct class in order to invert it), and therefore does not effectively remove private information from the representation.

To avoid this issue, in this work we rely on a differentiable proxy for the mutual information between the output $\hat{z}$ of a privacy head and the private labels, similarly to \cite{irene}. For a given head output $\hat{z}_i$, this proxy of mutual information is computed as
\begin{equation}
    \label{eq:mi}
\mathcal{I}_z(\hat{z}_i) =\sum^{|\mathcal{Z}|}_{j=1} \sum^{|\mathcal{Z}|}_{k=1} \tilde{p}(\hat{z}_i=j, z = k)\log\left[\frac{\tilde{p}(\hat{z}_i=j, z = k)}{\tilde{p}(\hat{z}_i=j)\tilde{p}(z = k)}\right],
\end{equation}
being, across the minibatch:
\begin{itemize}
    \item $\tilde{p}(\hat{z}_i=j)$, the empirical frequency of private class $j$,
    \item $\tilde{p}(z = k)$, the mean of the $k^{\text{th}}$ softmaxed output of the privacy head,
    \item $\tilde{p}(\hat{z}_i=j, z = k)$, the mean of the $k^{\text{th}}$ softmaxed output of the privacy head when the true label is $j$.
\end{itemize}

Our optimization function can thus be defined as
\begin{equation}
    \label{eq:MI_optimization}
    \mathcal{J}=\mathcal{L}_y(\hat{y}) + \sum_{i=1}^B \gamma_i \mathcal{I}_z(\hat{z}_i),
\end{equation}

where $\{\gamma_i\}_{i=1}^B = \gamma \  \forall i$, and $\gamma$ is a fundamental hyperparameter controlling the strength of the regularization for each privacy head. In our experiment, we set $\gamma$ with a tuning procedure choosing the value bringing the lowest validation accuracy of the last privacy head. 
Contrary to maximizing $\mathcal{L}_z$, minimizing $\mathcal{I}_z$ forces the representation to discard information about the private attribute: in the limit $\mathcal{I}_z \to 0$, the optimal privacy classifier is forced to avoid exploiting the prior over $z$, and its accuracy is upper bounded by $\max_{z \in \mathcal{Z}} p(z)$ (which reduces to $1/|\mathcal{Z}|$ only in the perfectly balanced case).
Our objective is therefore to obtain block outputs $\{h_i\}_{i=1}^B$ such that the privacy heads cannot extract meaningful private information from them.

In practice, we first pre-train the model with the objective $\mathcal{J}$ defined in~\eqref{eq:MI_optimization} for a fixed number of epochs. Specifically, for each mini-batch, we alternate between:
\begin{itemize}
    \item updating the privacy heads $\{g^{p}_{i}\}_{i=1}^B$ with the cross-entropy loss $\sum_{i=1}^B \mathcal{L}_z(\hat{z}_i)$ while keeping the backbone $f$ fixed;
    \item updating the backbone and task head by minimizing $\mathcal{J}$ while keeping all privacy heads fixed.
\end{itemize}
The same batchwise alternating scheme is reused whenever we fine-tune a pruned model, so that the privacy objective is enforced consistently throughout all training stages. 

\subsection{Designed pruning strategy}
\label{sec:pruning}
Building on the insights of \cite{10.1007/978-3-031-73013-9_25}, we know that depending on the task, and the private features at hand, it is possible to prune very heavily specific layers in the network without compromising performance on the main task, while other layers tend to be very sensitive to pruning. Therefore, we prune our network by gradually increasing a \emph{global} sparsity level $s$ applied to all blocks, within an upper bound on the accuracy of the privacy head connected to the last network's block, that we employ as a proxy for the private attribute flow.

We use $L_1$-norm structured pruning to maximize the impact on memory efficiency and energy consumption. Specifically, our pruning approach removes entire channels from layers based on their (normalized) $L_1$-norm values. Let us consider a convolutional layer with weight tensor $\mathbf{W}_{i} \in \mathbb{R}^{C_{\text{out}} \times C_{\text{in}} \times K_H \times K_W}$, where $c$ indexes channels, $j$ indexes input channels, and $(h,w)$ index spatial kernel dimensions. \\ For each output channel $c$ in a prunable layer, we compute the normalized $L_1$-norm
\[
\|\mathbf{W}_{i}[c, :, :, :]\|_1 = \frac{1}{C_{\text{in}} K_H K_W}\sum_{j,h,w} |\mathbf{W}_{i}[c,j,h,w]|,
\]
and then permanently remove the channels with the lowest norms by applying a structured mask $\mathcal{M}_i^{s} \in \{0,1\}^{C_{\text{out}}}$ where $\mathcal{M}_i^{s}[c] = 1$ if channel $c$ is among the top-$(1-s) \cdot C_{\text{out}}$ channels of its layer by normalized $L_1$-norm, and $\mathcal{M}_i^{s}[c] = 0$ otherwise, with $s$ being the target sparsity level. 

This structured approach offers significant practical advantages over unstructured pruning methods:
\begin{itemize}
    \item \textit{Memory-wise}, removing entire channels reduces the actual tensor dimensions from $\mathbb{R}^{C_{\text{out}} \times C_{\text{in}} \times K_H \times K_W}$ to $\mathbb{R}^{(1-s) \cdot C_{\text{out}} \times C_{\text{in}} \times K_H \times K_W}$, leading to genuine memory savings during both training and inference.
\item \textit{Computationally}, eliminating output channels reduces the number of convolution operations from $O(C_{\text{out}} \cdot C_{\text{in}} \cdot K_H \cdot K_W \cdot H \cdot W)$ to $O((1-s) \cdot C_{\text{out}} \cdot C_{\text{in}} \cdot K_H \cdot K_W \cdot H \cdot W)$, resulting in measurable speedup proportional to the sparsity level. 
\end{itemize}
 These benefits in model size and energy consumption are shown in Sec.~\ref{sec:4_experiments}.
\\ 
\noindent \textbf{Working with blocks.} While we ideally could define and prune channels at the output of each and every layer of a model, this can, in practice, result in too long an optimization process and in architectures that are difficult to deploy efficiently. Furthermore, in models such as ResNets \cite{resnet} that have skip connections, information can bypass some layers. For these reasons, we group layers into \emph{blocks}. The structured pruning scheme described above is applied channel-wise across all blocks of the network, while the privacy objective is enforced at the end of each block via its corresponding privacy head.

\section{Experiments}

\label{sec:4_experiments}
\subsection{Setup}
\label{subsec:imp}
\textbf{Datasets.} We perform experiments on both syntethic and real-world datasets, evaluating CutClean performance with diverse private attributes. Specifically, we employ Corrupted-Cifar10 \cite{nam2020learning_lff} where the private attribute corresponds to image corruptions, Waterbirds \cite{Sagawa*2020Distributionally_groupDRO} where it is in the background, and CelebA \cite{celeba}, a face dataset that present multiple labels, with high-level sensitive attributes such as gender or ethnicity. Here, we consider two target attributes for CelebA, that is "blonde" and "heavy make-up". For all these datasets, we build custom versions balanced with respect to target and private attribute, as detailed in the Supplementary material.\\
\textbf{Data pre-processing and pruning details.}
For each model, we start from ImageNet-1k pre-trained weights. For each dataset, images are resized to $224\times224$, and we employ horizontal flips and random crops during training. For ResNet18 \cite{resnet}, we consider each residual stage as a pruning block. For the experiments with ViT-B16, we treat each transformer stage between downsampling operations as a block, and we prune the output channels of convolutional projections and feed-forward layers consistently within a block. The last privacy head threshold is defined according to the number of classes. Specifically, we set $P^{\text{threshold}} = 65\%$ for waterbirds and Celeb-A, with two target classes, and $P^{\text{threshold}} = 20\%$ for the syntethic dataset Corrupted-CIFAR10.\\
\textbf{Training details.} We employ SGD with momentum $0.9$ as optimizer, with a cosine learning-rate decay. The MI pretraining phase runs for $400$ epochs with an initial learning rate of $10^{-2}$, while the MI fine-tuning phase after pruning runs for $10$ epochs with an initial learning rate of $10^{-2}$. Weight decay is set to $10^{-4}$, and a batch size of $124$ is used for all the datasets. 

\subsection{Results}
\label{sec:results}
In this section, we present CutClean results on the selected datasets, reporting the best $\gamma$ selection procedure, and then evaluating the corresponding pruned model in terms of sparsity, target and privacy head test accuracy. \\
\textbf{Results on Celeb-A: Blonde target attribute.}
Table~\ref{tab:gamma_selection_celeba_blondhair_gender} shows the target accuracy and the last privacy head accuracy on the validation set for Celeb-A, considering the target attribute ``blonde", for different values of $\gamma$. We can notice how $\gamma =0$, which corresponds to our baseline model with no MI-aware training, brings a high validation accuracy for the last privacy head of $90.61\%$, confirming a clear privacy information flow through the model. On the other hand, $\gamma = 1000$ provides the lowest privacy head accuracy and is therefore selected for the rest of our pipeline. Table \ref{tab:CutClean_BLonde} summarizes the corresponding CutClean results. While preserving task accuracy, with a negligible reduction of less than $1.5 \%$ on the test accuracy, our method allows to obtain a significant reduction of the privacy information flow, with a minus $24.86 \%$ in the privacy head accuracy, further allowing to compress the model with a sparsity level of $60 \%$. 
\begin{table}[!t]
\centering
\caption{Target and last privacy head accuracy on the validation set for different values of $\gamma$ for Celeb-A "blonde" attribute.}
\label{tab:gamma_selection_celeba_blondhair_gender}
\begin{tabular}{rcc}
\toprule
$\gamma$ & \texttt{Target\ Accuracy (\%)} & \texttt{Last\ PH\ accuracy (\%)}  \\
\midrule
0               & $91.71 \pm 0.84$ & $90.61 \pm 0.45$ \\
0.01            & $92.17 \pm 0.79$ & $87.68 \pm 0.58$ \\
0.10            & $89.93 \pm 0.58$ & $74.13 \pm 1.69$ \\
1               & $89.15 \pm 1.85$ & $71.34 \pm 2.38$ \\
5               & $89.47 \pm 0.75$ & $71.15 \pm 0.67$ \\
10              & $89.61 \pm 0.26$ & $69 \pm 0.66$ \\
50              & $89.47 \pm 0.64$ & $65.75 \pm 2.57$ \\
100             & $90.02 \pm 0.39$ & $64.06 \pm 1.88$ \\
500             & $89.29 \pm 1.36$ & $61.45 \pm 1.82$ \\
$\gamma^\star=1000$ & $89.65 \pm 0.71$ & \textbf{61.13 $\pm$ 1.46} \\
\bottomrule
\end{tabular}
\end{table}
\begin{table}[!t]
\centering
\caption{Target accuracy, last privacy head accuracy on the test set and sparsity level on Celeb-A "blonde" attribute.}
\label{tab:CutClean_BLonde}
\small
\resizebox{0.9\textwidth}{!}{%
\begin{tabular}{@{}lrrrr@{}}
\toprule
Model & Target Accuracy (\%) & Last PH Accuracy (\%) & Sparsity (\%) \\
\midrule
Baseline & $91.81 \pm 0.85$ & $87.59 \pm 1.14$ & 0 \\
CutClean & $90.32 \pm 0.64$ & $62.73 \pm 1.24$ & 60 \\
CutClean w/o fine-tuning & $88.84 \pm 1.16$ & $61.02 \pm 2.36$ & 5 \\
\bottomrule
\end{tabular}}
\end{table}
\\\textbf{Results on Celeb-A: Heavy make-up target attribute.}
Table \ref{tab:gamma_selection_celeba_heavymakeup_gender} summarizes the target and the last privacy head validation accuracy for the attribute "Heavy make-up" in Celeb-A with respect to different values of $\gamma$. Similarly to the previous attribute, a naive training ($\gamma=0$) brings a high accuracy in the last privacy head, while $\gamma=5$ allows to reduce privacy flow with a decreasing in the privacy head accuracy of almost $20\%$, and is therefore selected for the rest of our pipeline. Table \ref{tab:CutClean_heavy} shows the corresponding CutClean results. On the test set, our pruned model achieves a reduction of privacy head accuracy of $15.91 \%$, while preserving a good target accuracy of $73.11 \%$, with a final sparsity of $30\%$.
\begin{table}[!t]
\centering
\caption{{Target and last privacy head accuracy on the validation set for different values of $\gamma$ for Celeb-A "Heavy make-up" attribute.}}
\label{tab:gamma_selection_celeba_heavymakeup_gender}
\begin{tabular}{rcc}
\toprule
$\gamma$ & \texttt{Target\ Accuracy (\%)} & \texttt{Last\ PH\ accuracy (\%)}  \\
\midrule
0               & $74.07 \pm 3.46$ & $81.48 \pm 7.29$ \\
0.01            & $78.70 \pm 1.31$ & $85.19 \pm 7.29$ \\
0.10            & $74.07 \pm 5.71$ & $78.70 \pm 4.72$ \\
1               & $73.15 \pm 4.72$ & $76.85 \pm 1.31$ \\
$\gamma^\star=5$    & $75 \pm 3.93$    & \textbf{62.04 $\pm$ 14.76} \\
10              & $70.37 \pm 2.62$ & $63.89 \pm 3.93$ \\
50              & $67.59 \pm 1.31$ & $67.59 \pm 1.31$ \\
100             & $68.52 \pm 1.31$ & $64.81 \pm 7.97$ \\
500             & $64.81 \pm 5.71$ & $69.44 \pm 7.86$ \\
1000            & $70.37 \pm 4.72$ & $64.81 \pm 2.62$ \\
\bottomrule
\end{tabular}
\end{table}
\begin{table}[!t]
\centering
\caption{CutClean performance on Celeb-A, "Heavy make-up" attribute}
\label{tab:CutClean_heavy}
\small

\resizebox{0.9\textwidth}{!}{%
\begin{tabular}{@{}lrrrr@{}}
\toprule
Model & Target Accuracy (\%) & Last PH Accuracy (\%) & Sparsity (\%) \\
\midrule
Baseline & $77.65 \pm 4.76$ & $80.30 \pm 0.53$ & 0 \\
CutClean & $73.11 \pm 6.29$ & $64.39 \pm 4.98$ & 30 \\
CutClean w/o fine-tuning & $55.30 \pm 9.19$ & $59.47 \pm 7.56$ & 40 \\

\bottomrule
\end{tabular}}
\end{table}
\\\textbf{Results on Corrupted-CIFAR10.}
Table \ref{tab:gamma_selection_corrupted_cifar10} shows the results of our hyperparameter tuning procedure for selecting the best $\gamma$ on Corrupted-CIFAR10. Despite the different semantics of the private attribute, which here is an image corruption, without our MI-aware training procedure ($\gamma=0$), the privacy head accuracy on the validation set is around $64\%$, confirming privacy information flow through the network. Similarly to the "blonde" target attribute in Celeb-A, $\gamma=1000$ provides the lowest privacy head accuracy, which is close to random guess ($14.57\%$), and is employed in the rest of CutClean procedure. Table \ref{tab:CutClean_corrupted} shows the results on the test set for the pruned model. Here, we can reach very low accuracy on the privacy head ($13.33\%$), while maintaining good performances on the target classification task ($81.36\%$), and with a corresponding sparsity value of $10\%$. 
\begin{table}[!h]
\centering
\caption{{Target and last privacy head accuracy on the validation set for different values of $\gamma$ for Corrupted-CIFAR10.}}
\label{tab:gamma_selection_corrupted_cifar10}
\begin{tabular}{rcc}
\toprule
$\gamma$ & \texttt{Target\ Accuracy (\%)} & \texttt{Last\ PH\ accuracy (\%)}  \\
\midrule
0               & $83.89 \pm 0.27$ & $64.41 \pm 0.81$ \\
0.01            & $84.05 \pm 0.51$ & $63.92 \pm 0.84$ \\
0.10            & $82.07 \pm 0.59$ & $41.93 \pm 0.54$ \\
1               & $80.09 \pm 0.13$ & $25.44 \pm 0.79$ \\
5               & $79.53 \pm 0.58$ & $19.67 \pm 0.51$ \\
10              & $78.11 \pm 0.49$ & $17.69 \pm 0.54$ \\
50              & $77.27 \pm 0.50$ & $16 \pm 1.50$ \\
100             & $76.84 \pm 0.78$ & $16.56 \pm 0.60$ \\
500             & $76.05 \pm 0.51$ & $15.97 \pm 0.66$ \\
$\gamma^\star=1000$ & $76.37 \pm 0.52$ & \textbf{14.57 $\pm$ 0.69} \\
\bottomrule
\end{tabular}
\end{table}
\begin{table}[!h]
\centering
\caption{Target accuracy, last privacy head accuracy on the test set and sparsity level on Corrupted-CIFAR10.}
\small
\label{tab:CutClean_corrupted}
\resizebox{0.9\textwidth}{!}{%
\begin{tabular}{@{}lrrrr@{}}
\toprule
Model & Target Accuracy (\%) & Last PH Accuracy (\%) & Sparsity (\%) \\
\midrule
Baseline & $84.23 \pm 0.49$ & $62.87 \pm 1.16$ & 0  \\
CutClean & $81.36 \pm 0.36$ & $13.33 \pm 0.50$ & 10 \\
CutClean w/o fine-tuning & $75.54 \pm 0.33$ & $14.59 \pm 0.97$ & 0 \\
\bottomrule
\end{tabular}}
\end{table}
\\\textbf{Results on Waterbirds.}
Table \ref{tab:gamma_selection_waterbirds} summarizes the target validation accuracy and the last privacy head accuracy at the varying of $\gamma$ for the waterbirds dataset, where the private attribute is the background. Here, with $\gamma=0$ the privacy information flow is very high, with a corresponding privacy head accuracy of $91.47\%$, while $\gamma=1$ allows to reduce this value up to $69.23\%$, and is used for the rest of CutClean pipeline. As reported in Table \ref{tab:cutclean_waterbirds}, with a negligible decrease of $2.21 \%$ in the target accuracy, our pruned models is capable of providing a reduction in the privacy head accuracy of $24.4 \%$, allowing to obtain a sparsity level of $20\%$.   
\begin{table}[!t]
\centering
\caption{Target and last privacy head accuracy on the validation set for different values of $\gamma$ for Waterbirds.}
\label{tab:gamma_selection_waterbirds}
\begin{tabular}{rcc}
\toprule
$\gamma$ & \texttt{Target\ Accuracy (\%)} & \texttt{Last\ PH\ accuracy (\%)}  \\
\midrule
0               & $90.31 \pm 0.07$ & $91.47 \pm 0.26$ \\
0.01            & $90.46 \pm 0.51$ & $90.36 \pm 0.93$ \\
0.10            & $89.46 \pm 0.61$ & $71.64 \pm 1.47$ \\
$\gamma^\star=1$    & $89.26 \pm 0.19$ & \textbf{69.23 $\pm$ 3.50} \\
5               & $88.20 \pm 0.84$ & $75.05 \pm 0.94$ \\
10              & $88 \pm 0.70$    & $77.51 \pm 2.86$ \\
50              & $88.76 \pm 0.31$ & $78.56 \pm 0.82$ \\
100             & $88 \pm 0.38$    & $80.27 \pm 1.30$ \\
500             & $87.20 \pm 0.65$ & $82.93 \pm 0.92$ \\
1000            & $87.25 \pm 0.62$ & $83.63 \pm 0.87$ \\
\bottomrule
\end{tabular}
\end{table}
\begin{table}[!t]
\centering
\caption{Target accuracy, last privacy head accuracy on the test set and sparsity level on waterbirds.}
\label{tab:cutclean_waterbirds}
\small
\resizebox{0.9\textwidth}{!}{%
\begin{tabular}{@{}lrrrr@{}}
\toprule
Model & Target Accuracy (\%) & Last PH Accuracy (\%) & Sparsity (\%)\\
\midrule
Baseline & $88.76 \pm 0.51$ & $88.96 \pm 0.61$ & 0 \\
CutClean & $86.55 \pm 0.75$ & $64.56 \pm 1.48$ & 20 \\
ClutClean w/o fine-tuning & $50.00 \pm 0.01$ & $50.00 \pm 0.01$ & 95 \\
\bottomrule
\end{tabular}}
\end{table}

\subsection{Ablations}
\label{sec:ablations}
In this Section, we perform ablation studies to evaluate the impact of the main components of our method. First, we explore vision transformer architectures, replicating our pipeline with a ViT-B16 model. Then, we measure the impact of fine-tuning pruned models and the last privacy head, prior to evaluation on the validation set. Finally, we provide the last privacy head and the target accuracy on the test set for our pruned models as a function of the complete grid of candidate sparsity levels.\\
\textbf{Results on Transformers.}
We here replicate the entire CutClean pipeline on the "blonde" target class for Celeb-A dataset, exploiting a ViTB-16 model. First, in Table \ref{tab:gamma_selection_celeba_blondhair_gender_vitb} we report the tuning experiment to select the best $\gamma$ value, confirming how a naive training procedure shows a significant leakage for the private attribute, while our procedure with $\gamma=50$ provides the best reduction in the last privacy head accuracy. Then, we employ this value of $\gamma$ for performing CutClean's pipeline, with results summarized in Table \ref{tab:CutClean_Vit}. Our results are similar to the ones obtained with ResNet18, allowing to obtain a sparsity level of $50\%$, with a corresponding reduction in the last privacy head accuracy of $25.11\%$, and a reduction in target accuracy of only $5.48\%$.
\begin{table}[!h]
\centering
\caption{{Target and last privacy head accuracy on the validation set for different values of $\gamma$ for Celeb-A "blonde" attribute, employing a ViTb-16 model.}}
\label{tab:gamma_selection_celeba_blondhair_gender_vitb}
\begin{tabular}{rcc}
\toprule
$\gamma$ & \texttt{Target\ Accuracy (\%)} & \texttt{Last\ PH\ accuracy (\%)}  \\
\midrule
0               & $90.20 \pm 1.60$ & $86.86 \pm 0.39$ \\
0.01            & $89.38 \pm 0.97$ & $86.08 \pm 1.04$ \\
0.10            & $87.13 \pm 1.19$ & $76.28 \pm 2.69$ \\
1               & $87.96 \pm 0.26$ & $73.58 \pm 1.80$ \\
5               & $87.41 \pm 1.13$ & $76.42 \pm 2.64$ \\
10              & $87.59 \pm 1.92$ & $70.42 \pm 1.72$ \\
$\gamma^\star=50$   & $86.90 \pm 2.22$ & \textbf{68.68 $\pm$ 4.13} \\
100             & $84.80 \pm 1.65$ & $70.92 \pm 2.27$ \\
500             & $83.52 \pm 0.79$ & $71.84 \pm 1.18$ \\
1000            & $82.60 \pm 0.51$ & $73.08 \pm 1.58$ \\
\bottomrule
\end{tabular}
\end{table}
\begin{table}[!h]
\centering
\caption{Target accuracy, last privacy head accuracy on the test set and sparsity level on Celeb-A "blonde" attribute, employing a ViTb-16 model.}
\small
\label{tab:CutClean_Vit}
\resizebox{0.9\textwidth}{!}{%
\begin{tabular}{@{}lrrrr@{}}
\toprule
Model & Target Accuracy (\%) & Last PH Accuracy (\%) & Sparsity (\%)\\
\midrule
Baseline & $90.19 \pm 1.04$ & $86.90 \pm 1.91$ & 0 \\
Without retrain after pruning & $79.03 \pm 3.01$ & $64.02 \pm 4.02$ & 10 \\
With retrain after pruning & $84.72 \pm 2.14$ & $61.73 \pm 2.84$ & 50 \\
\bottomrule
\end{tabular}}
\end{table}
\begin{figure}[!t]
\caption{Target and privacy head accuracy on the test set for the investigated datasets for CutClean (Left) and CutClean without fine-tuning (Right). Results are averaged across three runs, standard deviations in the shaded area.}
  \centering
  \label{fig:complete}
\includegraphics[
      width=0.9\textwidth,
 ]{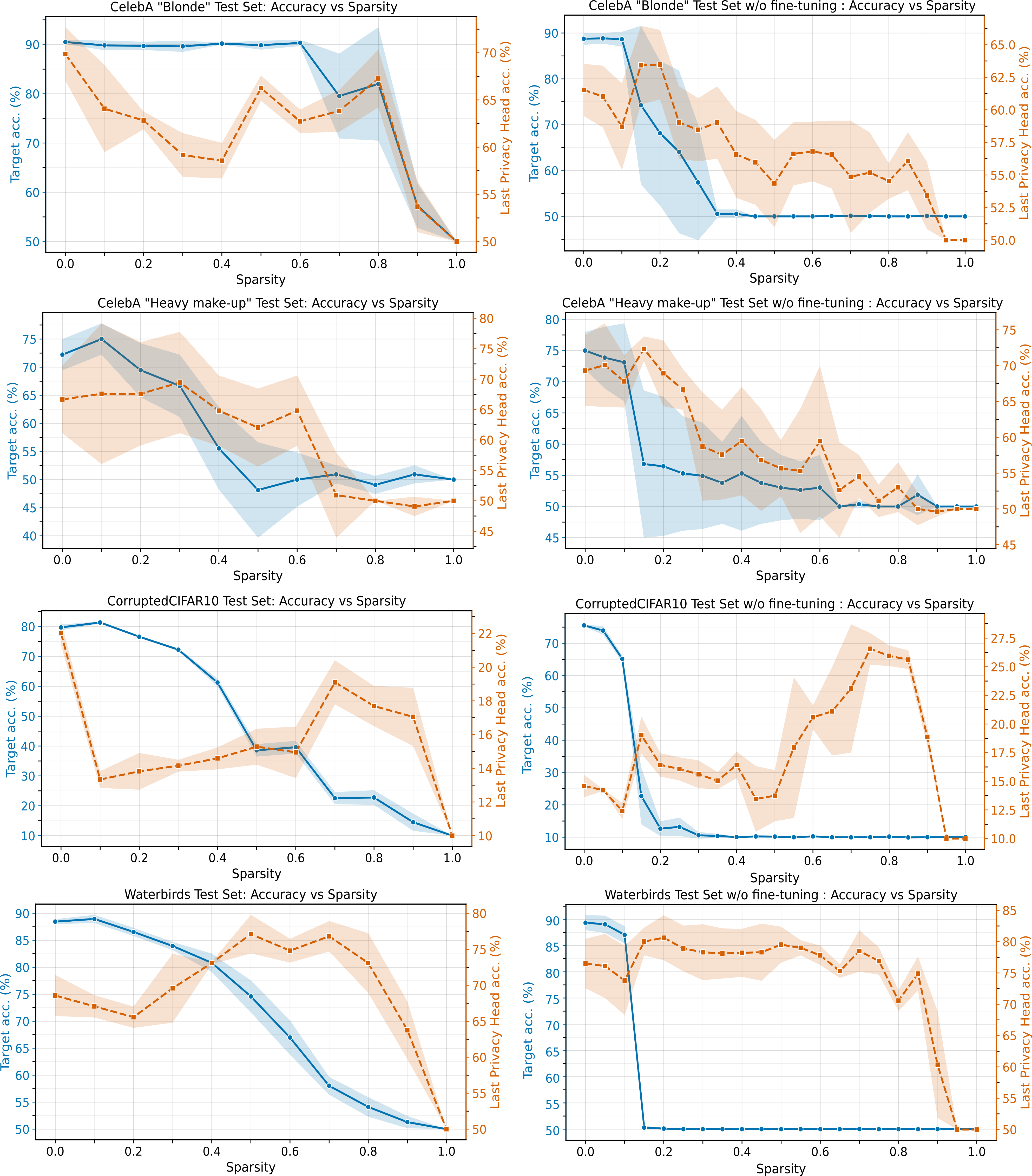}
  \end{figure}
\\\textbf{Impact of fine-tuning.}
We replicate our approach without fine-tuning the pruned models prior to evaluation on the validation set. The results reported in the last line of Tables~\ref{tab:CutClean_BLonde},
\ref{tab:CutClean_heavy},
\ref{tab:CutClean_corrupted},
and~\ref{tab:cutclean_waterbirds} confirm how fine-tuning allows to retain more the target classification performance of the original model. Furthermore, as within the admissible sparsity levels we select the one corresponding to the pruned models with the highest validation accuracy, even when the target performance can be partially recovered, we end-up selecting models with significantly lower sparsity values (e.g., for waterbirds and "blonde" target attribute in Celeb-A). Analogous results are obtained for the ViT-B16 model (last entry of Table \ref{tab:CutClean_Vit}), confirming the importance of fine-tuning in our method. \\
\noindent \textbf{Performance at different sparsity levels.}
In Fig.~\ref{fig:complete} we report the target and the last privacy head accuracy on the test set, for our complete set of candidate sparsity levels. Again, fine-tuning proves to better maintain the target classification performance for all the datasets, allowing to prune more aggressively before obtaining a severe reduction (for instance, up to $60\%$ of sparsity for the "blonde" target attribute in Celeb-A). Generally, higher level of sparsity values force the pruned models towards random guess, though the last privacy head accuracy shows a slighly different behavior, being capable of recovering private information for higher level of sparsities, as in the case of Celeb-A concerning the "blonde" target attribute. Without fine-tuning, moderate sparsity values are already sufficient to completely destroy the target classification performance, leading to random guess. This value can be as low as $0.15$ for waterbirds, or higher as in Celeb-A "blonde" target attribute, where it is $0.35$.
\section{Conclusion}

\label{sec:5_conclusion}
Despite the wide-spread adoption of deep neural networks, privacy leakage is still a major concern, potentially impacting real-world applications. On the other hand, it is known that deep neural networks are computationally expensive, with pruning techniques generally proposed to improve efficiency while maintaining performance on the target task. In this work, we tackle both issues proposing CutClean, a privacy-aware pruning method capable of significantly reducing the private information flow in a neural network. Exploiting external classifiers connected at each block of the network, that we refer to as privacy heads, we obtain a proxy of the privacy information flow at each block. Spanning over a set of candidate sparsity levels, we prune the network with our custom procedure, considering each level as admissible, if the last privacy head accuracy is below a fixed threshold, set according to the number of target classes, finally selecting the model that performs best on the validation set, in terms of target accuracy. Our results on four datasets confirm that with naive training there is a significant private information leakage, successfully reduced in the pruned models obtained through the proposed CutClean method, with consistent findings on vision transformer architectures. 

\section*{Acknowledgements}
This work is supported by Hi! PARIS and ANR/France 2030 program (ANR-23-IACL-0005).

\bibliographystyle{splncs04}
\bibliography{main}

\end{document}